\documentclass[11pt]{article}

\usepackage[margin=1in]{geometry}
\usepackage[T1]{fontenc}
\usepackage{lmodern}
\usepackage{microtype}
\usepackage{booktabs,tabularx,array}
\usepackage{enumitem}
\usepackage{float}
\usepackage{xcolor}
\usepackage{graphicx}
\usepackage[round,authoryear]{natbib}
\usepackage{hyperref}
\usepackage[nameinlink,noabbrev]{cleveref}
\usepackage{xcolor}
\usepackage{color}
\usepackage{soul}

\definecolor{linkblue}{HTML}{155F85}
\hypersetup{
  colorlinks=true,
  linkcolor=linkblue,
  citecolor=linkblue,
  urlcolor=linkblue,
  pdfauthor={Zuojun Max Shen; Yuan Qu; Pujun Zhang; Anbang Liu; Yunhao Liang},
  pdftitle={Enactive AI}
}

\setlist{nosep,leftmargin=*}
\newcolumntype{Y}{>{\raggedright\arraybackslash}X}
\newcolumntype{C}{>{\centering\arraybackslash}X}

\title{\textbf{Enactive Artificial Intelligence:
A Decision-Centric Architecture for Complex Systems}}
\author{
  \textbf{Zuo-Jun Max Shen} \quad \textbf{Yuan Qu} \quad \textbf{Pujun Zhang}\\[1mm]
  \textbf{Anbang Liu} \quad \textbf{Yunhao Liang} \quad \textbf{Feier Yan}\\[2mm]
  \small The University of Hong Kong \& OptiMax AI Limited, Hong Kong SAR, China\\[1mm]
  \small \href{mailto:maxshen@hku.hk}{maxshen@hku.hk} \quad
  \href{mailto:yuanqu@hku.hk}{yuanqu@hku.hk} \quad
  \href{mailto:pjzhang@hku.hk}{pjzhang@hku.hk}\\
  \small \href{mailto:anbang@hku.hk}{anbang@hku.hk} \quad
  \href{mailto:yunhao8@hku.hk}{yunhao8@hku.hk} \quad
  \href{mailto:yanfer@hku.hk}{yanfer@connect.hku.hk}
}

\begin{document}
\maketitle

\begin{abstract}
As artificial intelligence (AI) continues to evolve and mature, recent AI practices have moved beyond large language models (LLMs) and text or image generation tasks, increasingly integrating tools, agents, and harnesses to solve real business and industrial problems. However, the power of AI is not verified under these real-world complex systems for various reasons, considering reliability, feasibility, resilience, and responsibility requirements in real commercial and industrial operations. This study synthesizes adjacent research and introduces \emph{Enactive AI} as a conceptual framework for enterprise and industry reasoning, site-level decision support, and execution feedback. Four complementary roles organize the framework: an \emph{Organizational World} defines operations management logic and an organizational behavior world model behind an enterprise from a strategic-institutional horizon; a \emph{Site World} defines a physically bounded industrial optimization and execution world model from an operational-realization horizon; \emph{Schema Intelligence} provides the coupling mechanism between two world models to weave various AI applications via two models; and \emph{Enactive Decision Cycle} triggers the self-evolving dynamic process to update and audit the entire framework. By foregrounding decision intelligence in complex systems, Enactive AI expands the frontier of AI from model capability to system-aware action, opening new possibilities for scalable, governable, and socially valuable AI deployment. To show the practical value of the framework, we present three industrial case studies across major e-commerce platforms, manufacturing supply chains, and telecommunications service operations. Enactive AI points toward a future in which AI progress is measured not only by what models can generate or automate, but by how reliably intelligent systems can support consequential action, responsible governance, and durable social value in the complex systems that shape modern life, which we believe will define the next frontier of AI research for enterprise-level and industrial complex systems.

\end{abstract}


\section{Introduction}
\label{sec:intro}
The modern arc of artificial intelligence (AI) has been a steady widening of what AI systems can represent, generate, and execute. 
For decades, AI systems were built around well-specified representative tasks: classification and prediction in machine learning~\citep{jordan2015machine}, pattern recognition through deep representation learning~\citep{lecun2015deep}, ranking for information retrieval~\citep{liu2009learning}, and reinforcement learning for control under defined objectives~\citep{mnih2015humanlevel}. 
Since 2017, the introduction of the Transformer architecture and the attention mechanism has ushered AI into a new era for generation tasks, which expanded the scale of sequence modeling across language and multimodal data~\citep{vaswani2017attention}. Since then, foundation models have reframed broad transfer across tasks~\citep{bommasani2021opportunities}, and large language models have made few-shot generalization~\citep{brown2020language}, instruction following~\citep{ouyang2022training}, and multimodal interaction~\citep{openai2024gpt4} visible in a single interface. 
Action is the next heroic moment of AI. Modern AI systems start to call tools~\citep{schick2023toolformer}, observe feedback while reasoning and acting~\citep{yao2023react}, and use memory, planning, and agent coordination to revise workflows over time~\citep{wang2024surveyagents}. Beyond academia, the same movement is visible in commercial coding agents such as Claude Code~\citep{anthropic2026claudecodeoverview} and OpenAI Codex~\citep{openai2026codex}, which organize AI assistance around codebases, tools, and iterative software-development workflows. AI is no longer only a predictive module or a text generator, but is becoming an action-bearing interface for real-world problem solving.

As these capabilities continue to evolve and mature, the boundary of AI deployment is shifting from task-level assistance to system-level decision participation. 
Recent evidence shows that AI is widely used to shape prioritization, allocation, timing, intervention, and recovery in public administration and risk assessment~\citep{dressel2018recidivism,robodebt2023report,rintakahila2024robodebt}, health and public-safety monitoring~\citep{lazer2014parable,wong2021sepsis}, automated driving~\citep{ntsb2019tempe}, and operations management in supply chains~\citep{cohen2026scmai,hu2024jdadvanced,shen2025jdfulfillment}. 
Across these settings, once AI outputs enter a decision system, they are translated through objectives, rights, resources, authority, physical constraints, execution delays, and feedback. 
Yet this deeper involvement in decision systems has also revealed a serious problem: when representation, feasibility, accountability, and feedback are not organized together, locally plausible AI-mediated decisions can become unreliable at the system level. The frontier problem is therefore whether AI-mediated action can remain grounded, constrained, accountable, and adaptive when its consequences unfold in complicated real-world applications and scenarios.

Such real-world scenarios, characterized by interdependent components, layered structures, nonlinear interactions, constraints, and feedback loops, are best understood as complex systems.
Classic work on complexity emphasized hierarchy, near decomposability, and the organization of interdependent parts~\citep{simon1962architecture}; system dynamics showed how well-intended local actions can create delayed and unintended consequences in social and managerial systems~\citep{sterman1994learning}. 
Recent studies have pushed the explanation of system complexity further, revealing deeper and more intricately entangled causes: complex systems exhibit aggregate behavior that cannot be reduced to isolated components~\citep{ladyman2013complex}; networked interdependence can amplify risk across scales~\citep{helbing2013globally}; and machine behavior is shaped by the social, economic, and institutional settings in which machines act~\citep{rahwan2019machine,selbst2019fairness}. 
From a modern decision science perspective, these cause explanations reveal that decision context is distributed across objectives, policies, data pipelines, physical resources, institutions, and human actors; system state is partially observed and continuously evolving; actions are coupled through hard constraints and multi-agent responses; and feedback often arrives late, distorted, or endogenous. 
In such systems, a local intervention may reshape the conditions under which later decisions are made, while a local disturbance may propagate through resources, processes, organizations, and control structures before its full consequences become visible~\citep{perrow1984normal,leveson2012engineering}.
These features of complex systems raise an important question: how to build a well-organized framework through which AI can be effectively embedded and used within the system?


Faced with the above challenge, this paper proposes \emph{Enactive AI} as a decision-centric architecture for AI-mediated decision-making in complex systems, where AI outputs enter decisions that reshape operational states, propagate through interdependent relations, and carry organizational accountability.
Solving this problem requires capabilities that current AI systems only partially provide: 
\begin{itemize}
    \item \emph{System-grounded forecasting}, so that forecasts are formed from temporal evolution, dependency structure, resource state, and physical or digital operating conditions rather than from context alone; 
    \item \emph{Consequence-grounded decision-making}, so that system dynamics are translated into feasible interventions under objectives, constraints, authority, and responsibility rather than left as descriptive simulation; and 
    \item \emph{System-wise judgment}, so that an AI-mediated decision is evaluated by its contribution to the coupled decision system after human responses, execution delays, operational constraints, and feedback are included. 
\end{itemize}
Enactive AI equips these capabilities through three core components and a cycle. The \emph{Organizational World} defines a world model of the operations management logic and organizational behaviors from a strategic-institutional horizon, which contains intent, objectives, policies, data and evidence, responsibility boundaries, and managerial semantics. 
The \emph{Site World} defines a physically bounded industrial execution world model from an operational-realization horizon, which involves operational state, physical or digital resources, constraints, executable actions, execution systems, and feedback signals. 
\emph{Schema Intelligence} provides the coupling mechanism between two world models, translating organizational intent into AI-ready context, optimization-ready constraints, and execution-ready state-action representations, while transferring operational feedback back into organizational meaning.
The \emph{Enactive Decision Cycle} is the dynamic form of the architecture, which maintains reasoning, intent, feasibility, governance, consequence, and learning in motion as the system changes. Topologically, these components connect organizational purpose to operational dependencies, flows, bottlenecks, and propagation pathways, so that local interventions can be understood through their system-wide effects. Temporally, they keep state evolution, execution windows, delayed consequences, feedback, and revision inside a continuous decision process. In this sense, Enactive AI treats decision intelligence as a system property, making AI-mediated action purposeful, feasible, accountable, and adaptive in the organizational and operational world whose future it helps shape.


\begin{figure}
    \centering
    \includegraphics[width=0.9\linewidth]{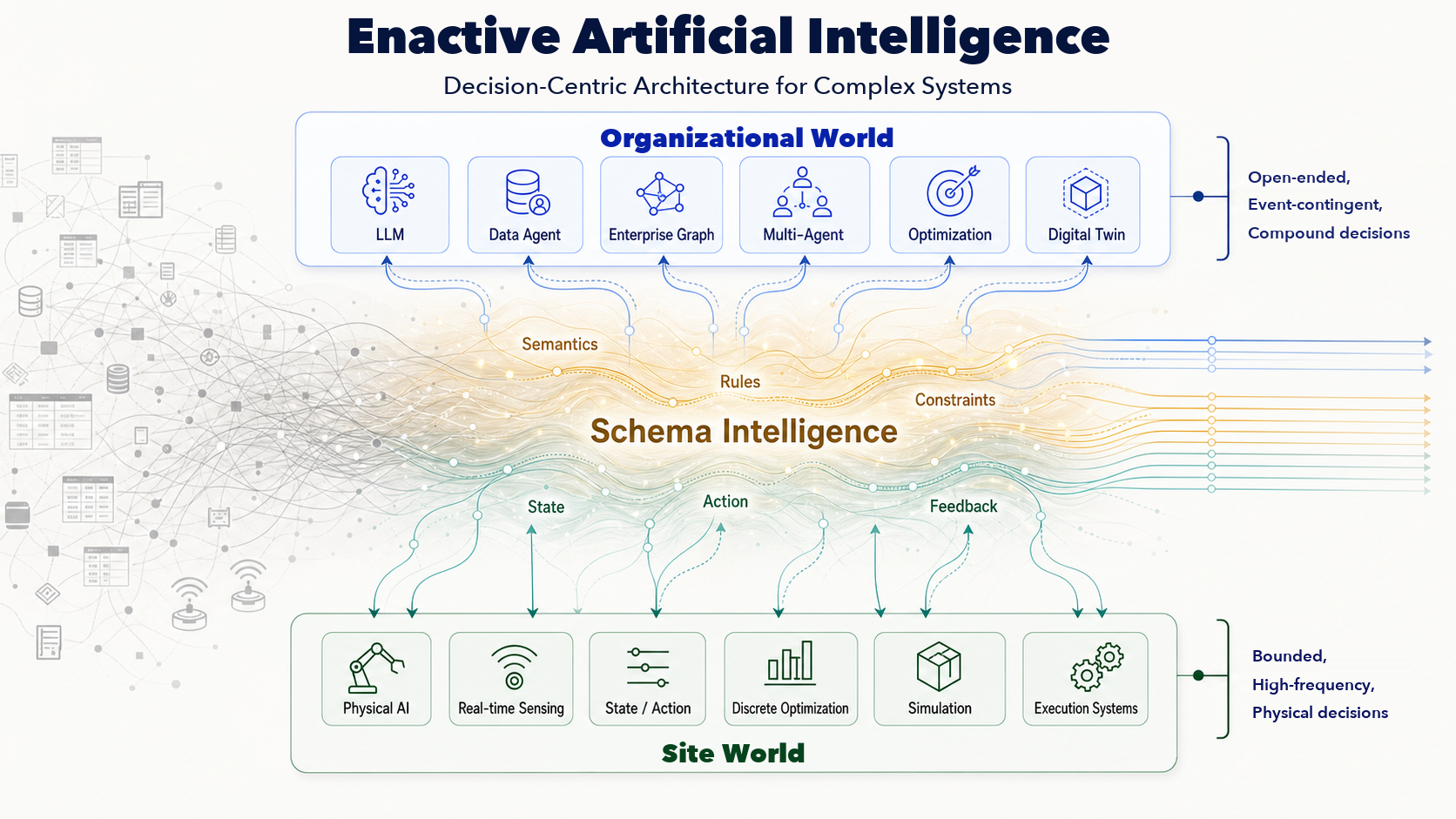}
    \caption{Enactive AI Architecture}
    \label{fig:EnactiveAIArchitecture}
\end{figure}

This architecture places Enactive AI in dialogue with a broader family of action-oriented AI ideas, and extends that view into complex organizational and operational decision systems. 
Embodied AI foregrounds perception-action coupling in physical environments~\citep{brooks1991representation,varela1991embodied}; world-model research learns environment dynamics for imagination, planning, and control~\citep{ha2018worldmodels,hafner2020dreamer,hafner2023dreamerv3,lecun2022path}; and existing enactive AI research asks how intelligence emerges from experience, autonomy, embodiment, and the inseparability of action and perception~\citep{froese2009enactive,rafiee2026toward}.
Our Enactive AI architecture, as developed here, carries the action-in-world insight to complex organizational and operational decision systems. Its focal object is no longer the individual agent, the learned environment model, or the simulator, but the decision system through which organizational purpose, operational state, feasible action, responsibility, consequence, and learning are continuously coupled.

The Enactive AI architecture brings a frontier complex-systems perspective to AI research and design. It directs attention to the structures through which organizations define value, impose constraints, authorize action, absorb consequences, and learn from change. In this view, intelligence is measured not only by what AI can generate or predict, but by the reliability of situated action: whether AI can preserve meaning, feasibility, accountability, and adaptation as decisions unfold through complex systems.
The contributions of this paper can be summarized as:
\begin{itemize}
    \item First, we point out AI-mediated decision-making in complex systems as a frontier problem that follows from AI's movement from representation and generation toward action. 
    \item Second, we articulate Enactive AI as a decision-system perspective for understanding how AI can support situated action under distributed context, constrained action, coupled consequences, and feedback-based evaluation. We develop an architectural expression of this perspective through Organizational World, Site World, Schema Intelligence, and the Enactive Decision Cycle. 
    \item Third, we outline a research agenda for evaluating, governing, and extending AI systems whose purpose is not merely to describe domains or complete workflows, but to participate responsibly in changing systems.
\end{itemize}



Amid the accelerating wave of AI development, decision intelligence in complex systems is becoming a defining frontier for the field. Enactive AI offers a perspective for this frontier: it asks AI research to move beyond stronger models and richer workflows toward the architectures that make intelligent action purposeful, feasible, accountable, and adaptive in the systems where consequences unfold. By bringing system-aware decision-making to the center of AI research and design, Enactive AI broadens the agenda of AI itself: it links capability with consequence, scale with governance, and deployment with durable social value. Its promise is not only to make AI more powerful, but to make AI more useful where society most needs intelligence: in the complex systems that allocate resources, coordinate work, manage risk, and shape collective futures.


\section{Development Review of AI and Complex Systems}
Over the past decade, AI has advanced at an exceptional pace, striding beyond conventional boundaries of prediction and classification toward generation and even action-oriented intelligence by utilizing tools, agents, memory, and harnesses.
However, the road to the next frontier for AI, decision-making intelligence in real-world complex systems, is crumpled.
In this section, we will review relevant research progress from both the evolution of AI capabilities and the deepening theoretical understanding of complex systems. 

\subsection{The Evolution of AI Capabilities}
For much of its early history, AI relied heavily on manually encoded knowledge, and there was little understanding of common-sense knowledge representation~\citep{mccarthy1987generality,feigenbaum1984fifth}. Early symbolic AI was developed around the idea that intelligence could be captured through explicit representations, rules, and search procedures~\citep{newell1976computer}. 
In the 1990s, the emergence of statistical machine learning shifted the focus from specifying domain knowledge explicitly to learning patterns from data. Data-mining and machine-learning systems learned mappings between observations and outcomes through statistical regularities in data~\citep{cortes1995support,breiman1996bagging,freund1997decision}. This paradigm supported a wide range of applications, including classification, prediction, recommendation, and decision support, where models could infer useful relationships without requiring complete human specification of the underlying rules~\citep{mitchell1997does,bishop2006pattern}.
From the mid-2000s onward, the rise of deep learning substantially improved how these systems learned representations from raw data~\citep{hinton2006reducing}. Deep neural networks could discover abstract patterns through multiple layers of representation learning, leading to major advances in perception, speech recognition, and sequential decision-making~\citep{lecun2015deep}. For instance, convolutional neural networks (CNNs) learned visual features directly from pixels and achieved breakthrough performance in image recognition~\citep{krizhevsky2012imagenet}, while deep reinforcement-learning systems learned control policies from high-dimensional observations through interaction with environments~\citep{mnih2015human}. Despite these advances, AI systems at this stage were still largely organized around task-specific objectives: a model was trained for a particular problem, evaluated under a predefined criterion, and deployed within a bounded environment.

In the late 2010s, advances in representation learning and foundation models brought AI systems into a new era. The introduction of the Transformer architecture provided a scalable mechanism for modeling long-range dependencies in sequential data, enabling large-scale pretraining across language and multimodal information~\citep{vaswani2017attention}. Building on this architecture, pretrained language models demonstrated that a single model could acquire broad linguistic knowledge and transfer it across a diverse range of downstream tasks. BERT showed that large-scale bidirectional pretraining could substantially improve language understanding across multiple benchmarks~\citep{devlin2019bert}, while GPT-3 demonstrated that sufficiently scaled language models could perform a wide variety of tasks through in-context learning without task-specific parameter updates~\citep{brown2020language}. From then on, the phrase large language model (LLM) is widely accepted due to the crazy size of the parameters in the pretrained models.
This trend was further strengthened by instruction tuning and alignment methods, which enabled models to better interpret human intentions and respond to open-ended requests~\citep{ouyang2022training}. The resulting foundation model paradigm represents a shift from developing separate models for individual tasks toward building general-purpose models that provide reusable representations and interfaces for a broad range of applications~\citep{bommasani2021opportunities,openai2024gpt4}. 
Whether in academia or industry, AI systems are no longer viewed merely as specialized predictive components, but largely as general problem-solving interfaces capable of interpreting information, synthesizing knowledge, and supporting interactions across previously separated domains.

Yet language generation was only the first visible expression of this broader change. As LLMs grew into massive numbers of capable interfaces for interpreting instructions and generating responses, academic and industrial vanguards started investigating their interactions with external tools, environments, and computational resources. Retrieval-augmented generation (RAG) linked LLMs to external knowledge sources to access information outside internal parameters for knowledge-intensive tasks~\citep{lewis2020retrieval}, and later systems learned to call calculators, search engines, translation services, question-answering systems, and other software via structured application programming interfaces~\citep{schick2023toolformer}. Building on these developments, agentic systems extended AI from generating responses to pursuing goals through sequences of actions. Early frameworks including ReAct combined reasoning traces with external actions to let LLMs alternate between reasoning and acting within interactive environments~\citep{yao2023react}, while other lines of research examined agents deployed in web environments, software engineering tasks and embodied settings, requiring systems to plan, engage with external states and iteratively improve their actions~\citep{yao2022webshop,yang2024swe}. 
Collectively, these advances represent a shift from AI systems focused chiefly on information generation toward systems that take part in workflows by orchestrating tools, carrying out multi-step procedures and adapting in response to environmental feedback.

As agentic capabilities matured, the focus of AI development rapidly shifted from individual agents toward larger systems that organize multiple capabilities, tools, and interactions into persistent workflows. Multi-agent systems extend the capability of a single agent by enlarging the context window size and enabling specialized agents to assume different roles, exchange information, coordinate intermediate tasks, and jointly solve complex problems~\citep{wu2024autogen, li2023camel}. Beyond reasoning by the agent itself, recent progress on surrounding infrastructure, often referred to in practice as agent harnesses, allows AI systems to operate more reliably over extended workflows~\citep{shinn2023reflexion, wang2024surveyagents}, with persistent memory, reusable skills, and standardized protocols provided. This transition is even more visible in commercial AI systems. Coding agents such as Claude Code and OpenAI Codex illustrate how LLMs can be integrated with terminals, software repositories, testing environments, and background execution processes to perform extended workflows rather than isolated responses~\citep{anthropic2026claudecodeworks,openai2026codex}. These developments mark a shift from AI systems that answer prompts toward AI systems that participate in workflows by interpreting goals, coordinating resources, executing actions, observing outcomes, and adapting subsequent behavior.

This movement from response generation toward action is beginning to change where AI is deployed. As AI systems acquire tool use, environmental feedback, planning, and workflow capabilities, AI-mediated decisions are gradually entering settings that were once governed mainly by human judgment, expert routines, or dedicated analytical systems.

\subsection{Complex Systems}
The expansion of AI capabilities has also shifted where AI participates. As AI systems move from controlled tasks toward real-world environments, the properties of those environments become central to the next stage of AI development. Many consequential domains are not simply large collections of tasks; they are complex systems in which behavior arises from interacting components, layered organization, evolving states, and feedback processes. A complex-systems perspective is therefore needed to explain why intelligence in such settings cannot be reduced to local prediction, isolated optimization, or single-step automation.


The intellectual roots of complex systems research can be traced to mid-twentieth-century attempts to move beyond reductionist explanation. Weaver distinguished problems of simplicity, disorganized complexity, and organized complexity, arguing that many biological, social, and technological problems involve many interrelated factors rather than either a few deterministic variables or merely statistical aggregates~\citep{weaver1948science}. General system theory and cybernetics further established a vocabulary for studying wholes, regulation, communication, and feedback across disciplinary boundaries~\citep{wiener1948cybernetics,ashby1956introduction,bertalanffy1968general}. This early systems tradition made interdependence and feedback central objects of inquiry, preparing the ground for later complexity science.

From the 1960s through the 1980s, a massive volume of research showed that system-level behavior cannot always be inferred from component-level analysis. Simon's account of hierarchy and near decomposability clarified why complex systems can often be analyzed through subsystems while still requiring attention to cross-level organization~\citep{simon1962architecture}. Anderson's claim that ``more is different" captured a broader shift toward emergence, where new regularities can appear at higher levels of organization that are not straightforward extensions of lower-level laws~\citep{anderson1972more}. In parallel, system dynamics emphasized feedback loops, accumulations, delays, and policy resistance in industrial and managerial systems~\citep{forrester1961industrial,sterman1994learning}, while nonlinear dynamics showed why deterministic systems can still display sensitive dependence and limited predictability~\citep{lorenz1963deterministic}. These developments shifted attention from static structure to dynamic behavior over time.

In the late twentieth century, complex adaptive systems became a unifying language for studying systems composed of interacting agents that learn, adapt, and co-evolve. Work associated with the Santa Fe Institute and related communities connected ideas from biology, economics, computation, and physics to explain how order, adaptation, path dependence, and novelty can emerge from decentralized interaction~\citep{holland1992complex,gellmann1995complexity,kauffman1993origins,arthur1999complexity}. Research on self-organization and criticality further showed how large-scale patterns and abrupt transitions may arise without centralized control~\citep{bak1987selforganized}. Complexity science thus moved from describing systems as interconnected structures toward understanding them as adaptive, evolving processes.

A subsequent wave of network science provided formal tools for examining how patterns of connection shape system behavior. Small-world networks, scale-free networks, and broader analyses of complex networks showed that topology influences diffusion, synchronization, robustness, vulnerability, and cascading effects across biological, technological, and social systems~\citep{watts1998collective,barabasi1999emergence,newman2003structure}. At the same time, work on high-risk and safety-critical sociotechnical systems emphasized that failures often arise not from a single faulty component, but from tight coupling, interactive complexity, organizational conditions, and inadequate control structures~\citep{perrow1984normal,leveson2012engineering}. These perspectives extended complex-systems thinking from natural and physical systems into the engineered, organizational, and institutional environments where modern AI is increasingly deployed.

Overall, the field's development suggests that complex systems should be understood not merely as large systems, but as systems whose behavior is shaped by interdependence, emergence, nonlinearity, adaptation, network structure, and feedback. Contemporary accounts similarly emphasize that complex systems exhibit aggregate behavior that cannot be reduced to isolated components and that their defining features vary across domains~\citep{mitchell2009complexity,ladyman2013complex}. For AI, the implication is direct: decisions in complex systems unfold through distributed context, constrained actions, coupled consequences, partial observation, delayed feedback, and adaptive responses. This accumulated body of work provides the conceptual foundation for examining AI decision-making in environments where actions alter the conditions under which subsequent decisions must be made.

\subsection{New Era of AI: Toward Decision Intelligence in Complex Systems}
\subsubsection{Emerging Applications of AI in Complex Systems}
The preceding review brings together two movements. AI systems have expanded from task-specific prediction toward general-purpose generation, tool use, agentic workflows, and action-oriented assistance. At the same time, complex-systems research shows that many real-world decision environments are organized through interdependence, feedback, evolving states, and cross-level constraints. The intersection of these two movements is now becoming increasingly visible: AI outputs are entering the routines through which complex systems are interpreted, prioritized, governed, and acted upon.

This shift can be seen across several concrete domains. In public administration and risk assessment, algorithmic systems help classify individuals, prioritize cases, and inform decisions about supervision, eligibility, and public obligations~\citep{dressel2018recidivism,robodebt2023report,rintakahila2024robodebt}. In health and public safety, predictive systems convert clinical records, behavioral traces, sensor streams, and environmental observations into signals for surveillance, triage, and intervention~\citep{lazer2014parable,wong2021sepsis}. In automated driving, AI-based perception and planning systems shape safety-critical trajectories under changing road conditions and human oversight~\citep{ntsb2019tempe}. In operations and supply chains, AI-enabled analytics support demand forecasting, replenishment, inventory allocation, fulfillment, and planning across distributed organizational networks~\citep{cohen2026scmai,hu2024jdadvanced,shen2025jdfulfillment}. These applications differ in domain and consequence, but they reveal a common expansion of AI's role from producing estimates about a system to informing the ranking, allocation, scheduling, planning, and execution of actions within it.

Across these domains, current AI applications are often organized around a recurring data--prediction--action pattern. Data or evidence are transformed into predictions or model outputs, and those outputs are then translated into recommendations, plans, or actions. Predictive models estimate quantities such as demand, risk, failure, or cost from observed data; rules, human decision makers, optimization models, or AI agents then use those estimates to rank alternatives, allocate resources, trigger interventions, or execute multi-step workflows. Prescriptive analytics and decision-focused learning have strengthened the connection between prediction and downstream optimization~\citep{bertsimas2020prescriptive,elmachtoub2022smart,donti2017taskbased}. The data--prediction--action chain is therefore a useful description of the dominant way AI enters many decision applications.

However, the literature has also documented recurring failures and difficulties that cannot be fully explained within this linear pattern. Studies of criminal-justice risk assessment question the accuracy and fairness of widely used tools, while the Robodebt case documents inaccurate automated debt notices, insufficient evidentiary grounding, legal invalidity, and the need for review, scrutiny, and audit of automated decisions~\citep{dressel2018recidivism,robodebt2023report,rintakahila2024robodebt}. Google Flu Trends is a well-known case of large errors in big-data prediction, and external validation of a widely implemented sepsis model found poor discrimination, poor calibration, missed cases, and alert fatigue~\citep{lazer2014parable,wong2021sepsis}. The NTSB's Tempe investigation identified limitations in hazard detection and path prediction together with inadequate safety risk assessment, operator oversight, automation-complacency controls, and public-road testing governance~\citep{ntsb2019tempe}. Recent operations-management research similarly identifies persistent challenges in data quality, model integration, governance, workforce adaptation, reliable data and infrastructure, transparent and explainable decision systems, and long-term human--AI collaboration in AI-enabled supply chains~\citep{cohen2026scmai}.

These problems reveal a deeper challenge for applying AI in complex systems. Once AI outputs enter consequential settings, they are mediated by evidentiary standards, rights, institutional procedures, shifting data-generating processes, intervention workflows, physical sensing, human supervision, safety-control structures, data infrastructure, organizational governance, and long-term human--AI collaboration. The central issue is therefore not only the performance of any single AI model, but the lack of a system-wise perspective that treats system context, feasible action, authority, execution, accountability, and feedback as constitutive parts of the decision process. 

\subsubsection{Capability Requirements for AI Applications in Complex Systems}
Viewed from a complex-systems perspective, reliable AI application requires more than local prediction, isolated optimization, or successful task execution. Because decisions unfold through interdependent components, changing states, constrained resources, institutional rules, human responses, and delayed feedback, AI systems must support three connected capabilities: \emph{system-grounded forecasting}, \emph{consequence-grounded decision-making}, and \emph{system-wise judgement}. These are not separate add-ons to the data--prediction--action chain. They specify how forecasting, action selection, and evaluation must be grounded in the same evolving decision system.

\paragraph{System-grounded forecasting}
The first capability requirement concerns forecasting. Most AI-assisted prediction remains grounded in statistical regularities observed in historical data. Traditional forecasting and supervised learning estimate a target variable from past observations, while large language models, trained through next-token prediction, produce explanations, forecasts, or action proposals conditioned on the textual context supplied in a decision setting~\citep{brown2020language}. Prescriptive analytics and decision-focused learning go further by connecting predictive models to downstream decision objectives or optimization losses~\citep{bertsimas2020prescriptive,elmachtoub2022smart,donti2017taskbased}. Nevertheless, the inputs and learning objectives in these approaches are still usually organized around a target variable, prompt, or bounded task rather than around the decision-relevant structure of the system. The conventional predictive question is therefore: given the observed data, what outcome is likely?

For decision-making in complex systems, this question is insufficient. A complex-system forecast must further ask what consequences may follow if a particular action is taken under the current system state. This question is action-conditioned: whether an intervention will change an outcome, how other actors will respond, whether resource bottlenecks will propagate, and whether relationships learned under previous conditions will remain valid after the intervention. Answering it requires forecasts to be formed from temporal evolution, dependency structure, resource state, and physical or digital operating conditions. In organizational and operational systems, relevant evidence is distributed across policies, contracts, databases, process definitions, analytical models, equipment states, and human expertise, with differences in semantics, authority, granularity, and update frequency. System-grounded forecasting therefore requires AI to construct decision-relevant representations of entities, dependencies, states, and dynamics rather than relying only on data-to-outcome prediction.

\paragraph{Consequence-grounded decision-making}
The second capability requirement concerns the pre-action movement from predicted consequence to feasible intervention. In current AI-assisted decision-making, predictions are commonly converted into actions through thresholds or business rules, supplied as parameters to optimization models that select a best action for a specified objective, or translated into plans and executable steps through agentic workflows. Human approval and safety checks are then often layered onto the resulting recommendation through surrounding organizational processes. These approaches can be effective when the decision problem, feasible set, objectives, and evaluation criteria have already been defined. Their central concern is accordingly which action best optimizes a specified objective.

Before action is taken in a complex system, however, the central question is not only which option has the highest score, but which intervention can and should be taken under the current objectives, constraints, authority, and responsibility. Candidate actions must be compared across multiple expected consequences, including cost, service, safety, resilience, and longer-term effects. They must also be checked against a feasible action space shaped by resource conditions, execution windows, policy constraints, decision rights, and execution readiness. Agentic execution does not by itself establish the right to act, just as optimization over encoded constraints does not establish that all system-relevant constraints and objectives have been represented~\citep{yao2023react,wang2024surveyagents,leveson2012engineering}. Consequence-grounded decision-making therefore requires AI to translate predicted system dynamics into feasible, authorized, and executable interventions.

\paragraph{System-wise judgement}
The third capability requirement concerns post-action and cross-cycle evaluation. Current AI systems are often evaluated by predictive accuracy, loss functions, benchmark performance, local objective value, or task completion. These measures are important, but they mainly assess the quality of a model output, an optimization result, or a bounded workflow. In complex systems, however, the relevant question is whether an AI-mediated decision improves the coupled decision system after the action has been interpreted, modified, executed, delayed, and fed back into future decisions.

System-wise judgement must therefore support system-level evaluation, outcome attribution, learning, and revision. Actions alter system states, influence other participants, and change the data on which later predictions are based~\citep{perdomo2020performative}. At the same time, the action recommended by a model may differ from the action approved, commanded, or ultimately realized because of human response, execution delay, partial implementation, operational failure, or adaptive behavior. When outcomes are delayed or jointly shaped by several interventions, an observed result cannot be attributed to the predictive model alone. It may reflect incomplete data, a missing system relationship, an inadequate consequence model, flawed judgement, or a deviation in execution. System-wise judgement therefore requires AI-mediated decisions to be evaluated by their contribution to the coupled decision system and used to revise the decision context as feedback arrives.

Together, these capability requirements show that current AI can connect data, prediction, and action, but reliable decision intelligence in complex systems requires a shared and evolving representation of the system in which evidence, state, feasible intervention, consequence, responsibility, execution, and feedback remain connected as conditions change. 

\section{The Architecture of Enactive AI}
\label{sec:framework}
This section introduces the architecture of Enactive AI and the concepts through which it is organized. It first provides an overview of the four functional elements: Schema Intelligence, the Organizational World, the Site World, and the Enactive Decision Cycle. The section then develops these elements in turn, before introducing System-Level Validity Governance and explaining how the architecture realizes system-grounded forecasting, consequence-grounded decision-making, and system-wise judgement.

\subsection{Architecture Overview}

\begin{figure}[H]
\centering
\includegraphics[width=\linewidth]{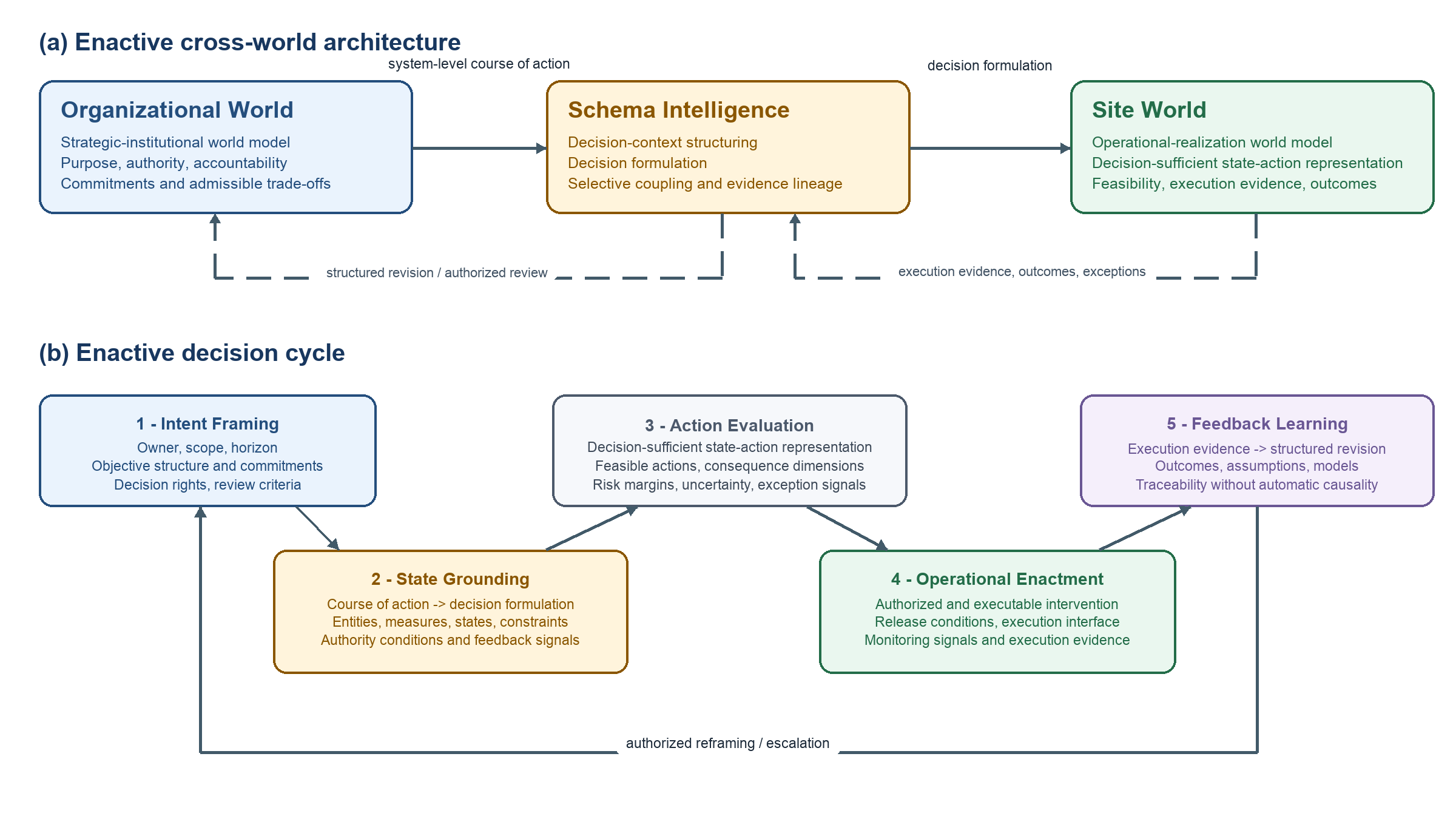}
\caption{Enactive AI as an operational decision system. Panel (a) distinguishes organizational framing, decision structuring, and bounded operational representation. Solid arrows show the translation from objectives and decision rights to an operational decision formulation; dashed arrows show how outcomes and exceptions support traceability and revision. Panel (b) presents an analytical decomposition of the recurring decision process.}
\label{fig:architecture}
\end{figure}

Figure~\ref{fig:architecture} presents Enactive AI as a decision architecture for complex systems. Its logic is organized through four functional elements: Schema Intelligence, the Organizational World, the Site World, and the Enactive Decision Cycle. 
\emph{Schema Intelligence} constructs and maintains the coupling mechanism that connects organizational meaning with operating entities, measures, states, actions, constraints, analytical models, and feedback, using semantic alignment, process models, data integration, optimization-model formulation, and analytical orchestration. 
The Organizational World defines the operations management logic and organizational
behaviors world model behind an enterprise from a strategic-institutional horizon, representing the managerial relations through which purpose, authority, accountability, commitments, and trade-offs are formed into system-level courses of action. 
The Site World defines a physically bounded industrial execution world model from an operational-
realization horizon, representing bounded operating subsystems in which those courses of action are tested against state, constraints, feasible actions, execution interfaces, and observed outcomes. 
The Enactive Decision Cycle is the self-evolved dynamic process that loops the entire framework, connecting framing, grounding, evaluation, enactment, and feedback.

The four elements are connected through a continuous relation between representation and action. Schema Intelligence organizes raw system inputs into decision-relevant elements that keep the two world models anchored in a compatible system context. From this context, the Organizational World reasons over purpose, authority, commitments, trade-offs, and cross-unit dependencies to form a \emph{system-level course of action}. Schema Intelligence then expresses that course of action as a \emph{decision formulation}, linking it to relevant objects, measures, states, constraints, authority conditions, model assumptions, feasible-action requirements, and feedback signals. The Site World grounds this formulation as a \emph{decision-sufficient state--action representation}, evaluates its feasibility under operating conditions, and returns execution evidence through Schema Intelligence to revise the decision context and inform both world models.

At this level, Enactive AI specifies how AI-mediated action can remain reliable as it enters systems whose conditions, constraints, and responses evolve. It supports three connected capabilities developed in the preceding sections: \emph{system-grounded forecasting}, \emph{consequence-grounded decision-making}, and \emph{system-wise judgement}. Forecasting is grounded in system state, dependency structure, and operating conditions; decision-making translates anticipated system dynamics into feasible, authorized, and executable interventions; judgement evaluates actions by their contribution to the coupled decision system and uses feedback to revise the decision context.

\subsection{Schema Intelligence}

In Enactive AI, Schema Intelligence is the capability for modeling an operational system as a decision-relevant topology structure. It identifies the entities that may participate in a decision and specifies the typed relations through which they constrain, enable, substitute for, or transmit action and consequence. Entities such as orders, facilities, suppliers, vehicles, machines, inventories, routes, contracts, policies, objectives, and model variables are not treated as isolated records. They are placed into a shared topology structure that connects operating entities, measures, flows, precedence relations, capacity-sharing relations, substitution options, common exposures, bottlenecks, authority conditions, and feedback signals. Through this schema, organizational meanings and operating facts acquire a common structural form for decision formulation, feasibility checking, consequence evaluation, execution, and revision.

The topology structure of an operational system gives decisions their system form. It determines which actions are feasible, which resources are substitutable, where bottlenecks appear, how disruptions travel, and how a local intervention may reshape downstream options. A local action can have very different implications depending on whether the affected entity is peripheral or central, whether it sits on a bridge or bottleneck, whether alternative paths or suppliers exist, and whether downstream commitments depend on it. In logistics and supply-chain settings, for example, routing feasibility, service reliability, disruption propagation, inventory pooling, supplier substitution, and recovery options are all shaped by network structure. Decision-making in such systems cannot be grounded only in local attributes or historical correlations; it must preserve the structural relations through which actions become feasible and consequences propagate.
\begin{figure}
    \centering
    \includegraphics[width=0.9\linewidth]{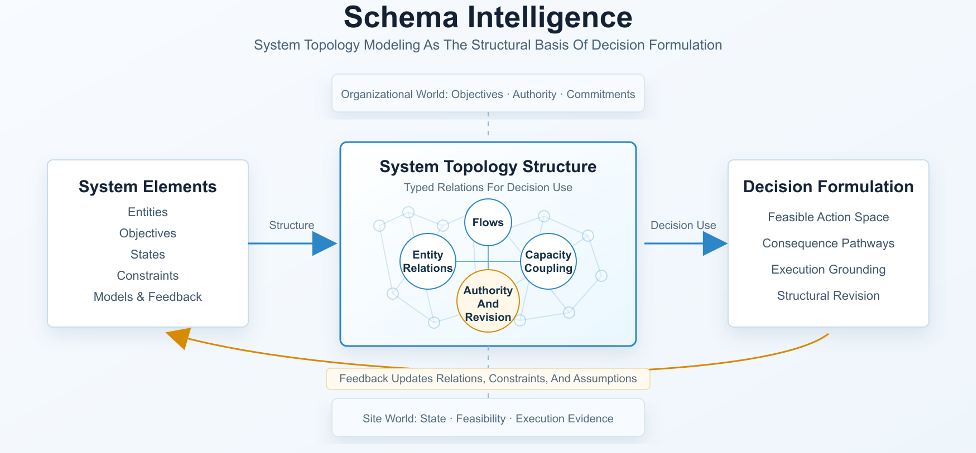}
    \caption{Schema Intelligence as system topology modeling for operational decision formulation.}
    \label{fig:schemaintelligence}
\end{figure}

Graph Optimization Foundation Model (GOFM) provides a concrete graph-native illustration of this principle \citep{liang2025graph}. In GOFM, a persistent weighted infrastructure graph is treated as a reusable decision asset rather than as an input to be reconstructed for each isolated optimization task. Structure-aware random walks convert the graph into a corpus of node-token trajectories that respect connectivity and edge weights. Progressive masked reconstruction then trains a Transformer to internalize the network's topology structure, distance geometry, reachability, bottlenecks, and subpath compatibility as a task-agnostic structural prior. At inference time, this prior is combined with task-specific constraints and objectives through lightweight constrained decoding, so that different routing and graph-analytic tasks can reuse the same learned structural backbone without retraining the entire model.

Schema Intelligence extends this structural view from graph optimization to broader operational decision systems. Its requirement is to preserve and reuse the relations that make a decision structurally meaningful across repeated decision episodes. The represented structure therefore expands from road-network nodes, edges, and weights to physical networks such as factories, warehouses, ports, and energy grids; process networks such as precedence relations, queues, handoffs, and workflow dependencies; supply networks such as sourcing links, substitutability, bill-of-material dependencies, inventory locations, and transportation lanes; organizational networks such as responsibility, approval, escalation, and accountability relations; and analytical networks such as data lineage, model assumptions, constraint definitions, and feedback signals. In this expanded setting, Schema Intelligence provides the reusable structure through which decision problems can be formulated, constrained, evaluated, enacted, and revised without treating each episode as an isolated task.

In a specific decision episode, this substrate is instantiated as a \emph{decision formulation}. A decision formulation is the operationally usable representation of a focal course of action. It specifies the decision boundary, the relevant entities and relation types, the feasible set induced by the system topology structure, current state and uncertainty, material constraints, objective-to-measure lineage, authority and admissibility conditions, model interfaces, candidate actions, consequence dimensions, and feedback signals. In graph terms, the formulation identifies which nodes, edges, weights, constraints, and relation types are material to the decision; in organizational terms, it identifies why those elements matter, who can authorize intervention, and how the outcome will be reviewed. Schema Intelligence therefore translates organizational intent into a structure-aware operational representation, rather than into a flat list of variables.

This formulation allows the two world models to reason over the same system structure from different horizons. The Organizational World uses it to evaluate whether a course of action remains coherent with purpose, commitments, trade-offs, decision rights, and cross-unit dependencies. The Site World uses it to evaluate whether that course can be grounded in current operating conditions, feasible actions, resource states, execution windows, bottlenecks, safety margins, and propagation pathways. A service-reliability decision, for instance, may need to connect customer commitments to orders, inventory, fulfillment capacity, depot locations, carrier availability, road segments, alternative routes, time windows, exception protocols, and review responsibilities. The importance of network structure is therefore not confined to optimization performance; it is what lets a managerial concern acquire operational content and lets an operational intervention remain accountable to managerial purpose.

The coupling maintained by Schema Intelligence is selective rather than exhaustive. Selective coupling means representing those cross-component, cross-horizon, or cross-system dependencies whose omission could change a course of action's feasibility, admissibility, consequences, or evaluation. This principle follows the logic of hierarchy and near decomposability: local data systems, analytical models, and operating routines can remain specialized, while dependencies that materially affect the focal problem remain visible \citep{simon1962architecture}. A complete enterprise ontology or full digital replica is neither necessary nor desirable. What is required is a decision-sufficient structural representation that keeps local decisions from being falsely treated as independent when their effects travel through shared resources, flows, commitments, or constraints.

Schema Intelligence also turns feedback into structural revision. Observed delay, infeasibility, deviation, service failure, inventory shortage, route disruption, or unexpected spillover may reveal that an edge weight has changed, a dependency was missing, a bottleneck was underestimated, a substitute relation no longer holds, a constraint has become active, or an authority condition is incomplete. The feedback loop therefore updates not only numerical parameters or model performance records, but also the relationships through which future decisions will be formulated. This lineage supports diagnosis, but it does not establish causality by itself; delayed and endogenous outcomes may remain ambiguous even when the decision history is explicit \citep{sterman1994learning}. Its value is to identify whether revision is needed in state representation, system topology structure, constraints, model assumptions, mappings, or organizational framing.

Schema Intelligence may draw on ontology and schema matching \citep{euzenat2013ontology}, semantic-layer and data-integration methods \citep{lenzerini2002data}, document retrieval and knowledge graphs \citep{lewis2020retrieval,hogan2021knowledge}, process mining and audit trails \citep{vanderaalst2016process}, network and graph representation learning, graph-native optimization foundation models \citep{liang2025graph}, optimization-model formulation \citep{ahmaditeshnizi2024optimus}, and agent orchestration \citep{yao2023react,wu2023autogen}, together with model registries and simulation interfaces for preserving model and execution lineage. Its boundary is equally important: it does not set organizational goals, generate system-level courses of action, or execute operational actions. It constructs and maintains the semantic and relational substrate for system topology modeling, through which goals, courses, states, actions, constraints, consequences, authority, and feedback can be reasoned over together.

\subsection{Organizational World}

In complex systems, an AI-mediated intervention acquires managerial significance only when it is embedded in organizational purpose, authority, and accountability. The Organizational World is the world model in which these managerial relations are represented and reasoned over. It articulates organizational purpose as a \emph{system-level course of action}: an integrated plan for pursuing that purpose through coordinated and mutually feasible actions across the system. Such a plan links objectives, commitments, trade-offs, coordination requirements, decision rights, and review criteria into a form that can guide downstream reasoning and execution.

The core modeling capability of the Organizational World is \emph{organizational course-of-action reasoning}. This capability is generative as well as predictive: it composes candidate system-level courses of action and projects their organizational consequences before they are translated into operational formulations. To do so, it represents recurrent relationships among objectives, commitments, authority, incentives, and cross-unit dependencies. These relationships allow it to reason about which courses of action are organizationally viable, where objective trade-offs or authority conflicts may arise, and how coordination burdens may propagate across units. The result is a system-level projection of organizational coherence and accountability that Schema Intelligence can operationalize and the Site World can ground in concrete operating conditions.

Governance is part of this reasoning rather than a later control layer. A course of action is not only a technically plausible response; it must also be admissible under decision rights, accountability relations, and escalation conditions. The Organizational World therefore represents how authority is distributed, how conflicting objectives are made visible, and how responsibility is attached to the release or revision of a system-level action. It supports governed action design without treating the AI system as the autonomous source of organizational goals or value judgments.

Taken together, these functions define the architectural departure of Enactive AI. Many AI applications begin after the objective, task boundary, and target variable have already been selected; the organizational problem has already been narrowed. In coupled organizations, this narrowing can create \emph{objective drift}, where local optimization preserves the assigned metric while displacing the broader purpose. The Organizational World keeps that broader purpose, its projected organizational consequences, and its admissible trade-offs visible while courses of action are generated and before operational formulation begins.

The Organizational World can draw on established methods such as multi-agent systems for distributed coordination \citep{wooldridge2009multiagent,shoham2009multiagent}, planning and optimization for course-of-action composition \citep{ghallab2004automated,boyd2004convex,bertsimas1997linear}, game theory and mechanism design for incentive alignment \citep{hurwicz1973mechanisms,nisan2007algorithmic,liang2025everyone}, and scenario planning, organization design, and workflow management for consequence projection and governed execution \citep{schoemaker1995scenario,galbraith1974organization,aghion1997formal,vanderAalst2004workflow}. These methods operate over organizational artifacts such as KPI definitions, contracts, policies, responsibility matrices, exception protocols, and audit trails. Semantic alignment of heterogeneous system records, documents, and model variables belongs to Schema Intelligence.

\subsection{Site World: operational state, feasible action, and execution.}

Within the Enactive AI architecture, the Site World is the operational world model that grounds a system-level course of action after Schema Intelligence has translated it into a decision formulation. It represents the current and anticipated operating conditions under which that formulation can be evaluated, qualified, and enacted. The Site World's basic unit is a \emph{decision-sufficient state--action representation}: the subset of state variables, resources, constraints, candidate actions, consequence estimates, and execution signals needed to determine whether the focal course of action can be realized and how it can be connected to execution.


A ``site'' is therefore a bounded operating subsystem rather than necessarily a single physical facility. Its boundary is determined by the entities, resources, dependencies, constraints, and execution interfaces that can materially affect the realization of a decision formulation. The Site World mediates two directions. Upward, it supplies operating evidence about feasibility, bottlenecks, risk, and realized outcomes that can qualify the formulation or trigger revision of the organizational course of action. Downward, it receives decision formulations that must be tested for technical feasibility and connected to execution. It preserves state-to-action, action-to-feasibility, action-to-consequence, commanded-to-realized-action, and action-to-outcome relationships.


A central design principle is decision sufficiency. The Site World retains the distinctions whose omission could change the technical feasibility, material consequences, risk, or comparison of candidate actions under the focal objective. Two operational situations may share the same representation when their differences have no material effect on the relevant action set or its consequences. A small change in equipment condition, resource availability, queue position, process status, or disruption exposure must be retained when it can make an action infeasible or alter its expected effects. Provenance, freshness, and uncertainty are therefore part of the representation, since an action that was feasible when evaluated may no longer be feasible when released.

In the Enactive Decision Process, the Site World primarily supports \emph{Action Evaluation}, \emph{Operational Enactment}, and \emph{Feedback Learning}. It converts operating evidence into a decision-relevant state and a set of candidate interventions. Depending on the setting, these interventions may include allocating capacity, resequencing work, replenishing or repositioning inventory, routing resources, expediting or deferring tasks, switching equipment, or activating recovery procedures. For each candidate, the Site World may represent cost, completion time, service impact, quality, resource consumption, congestion, operational risk, spillovers, recovery requirements, and uncertainty. Keeping these consequences distinct allows alternatives to be reconsidered when organizational priorities change and supports action-to-consequence evaluability.

The Site World also establishes the technical qualification of candidate actions. An intervention enters the technically feasible set only when it is compatible with the current operating state, resource capabilities, process dependencies, timing requirements, and applicable safety conditions. Organizational admissibility remains a separate condition connected through Schema Intelligence. A technically feasible action may still be withheld because of policies, contractual commitments, approval requirements, decision rights, or independent safeguards.

The Site World may be implemented through sensing systems, Internet-of-Things platforms, computer vision, transactional systems, forecasting, operations research models, simulation, digital twins, physical AI, robotics, controllers, and execution logs \citep{haller2017ssn,lecun2015deep,lim2021temporal,
bertsimas2020predictive,law2015simulation,kritzinger2018digital,nvidia2025cosmos}. These technologies may be combined with operational artifacts such as standard operating procedures, safety rules, dispatch policies, fallback procedures, exception protocols, maintenance rules, and recovery plans. Their configuration depends on the focal decision, the operating environment, and the available execution infrastructure.

Once an authorized intervention enters execution, the Site World maintains its connection to the action actually carried out and records delay, deviation, override, interruption, failure, and recovery. Expected consequences can then be compared with realized actions and observed outcomes. This comparison may update the operating-state estimate, consequence representation, feasibility conditions, or candidate-generation process, and may reveal a missing dependency that Schema Intelligence should revise.

The effective boundary of the Site World depends on state familiarity, evidence freshness, consequence uncertainty, candidate separation, risk margins, and the available decision window. When evidence is stale, the state is unfamiliar, candidate consequences are difficult to distinguish, or an action approaches a feasibility or safety boundary, the Site World should request additional evidence, deeper optimization or simulation, human review, escalation, or fallback. It complements transactional platforms, deterministic controllers, professional judgment, and independent safety mechanisms. Changes to organizational objectives, policies, or decision rights remain subject to a separate authorized organizational process.

\subsection{The Enactive Decision Cycle}
\label{sec:cycle}

The Enactive Decision Cycle is the temporal form of the architecture. It describes how Enactive AI sustains continuity among organizational purpose, decision context, feasible action, execution, and feedback as conditions change. The five stages are an analytical decomposition rather than a rigid workflow: a decision episode may pause, return to an earlier stage, or be escalated when evidence, authority, feasibility, or consequence estimates become unstable.

\noindent\textbf{1. Intent Framing.} Intent Framing is led by the Organizational World. It turns organizational purpose into a \emph{system-level course of action}: an integrated plan for pursuing a focal purpose through coordinated and mutually feasible actions across the system. The stage makes explicit the owner, scope, horizon, objective structure, admissible trade-offs, commitments, decision rights, coordination requirements, and review criteria. Its value is to keep the decision from collapsing too early into a local metric or narrow task boundary before the broader organizational consequences have been reasoned over.

\noindent\textbf{2. State Grounding.} State Grounding is led by Schema Intelligence. It instantiates the course of action as a \emph{decision formulation} by linking organizational objectives and commitments to relevant entities, measures, operating states, constraints, model assumptions, authority conditions, feasible-action requirements, and feedback signals. This stage draws on the shared semantic substrate prepared from distributed records, documents, process traces, models, logs, and rules, while incorporating evidence specific to the current decision episode. The result is a formulation that the Organizational World and the Site World can both interpret.

\noindent\textbf{3. Action Evaluation.} Action Evaluation is carried out in the Site World over the decision formulation maintained by Schema Intelligence. The formulation is grounded as a \emph{decision-sufficient state--action representation}, identifying the operating state, feasible action set, material constraints, consequence dimensions, risk margins, and uncertainty needed for the focal decision. Candidate interventions are assessed for technical feasibility and projected consequences while keeping material trade-offs visible. If feasibility is weak, evidence is stale, consequences cannot be distinguished, or an action approaches an operating or safety boundary, the stage returns an exception signal rather than forcing a recommendation.

\noindent\textbf{4. Operational Enactment.} Operational Enactment converts a qualified candidate into an authorized and executable intervention. Before release, the Site World revalidates technical feasibility against current operating conditions; the Organizational World supplies the admissibility, accountability, and decision-right conditions under which the intervention may be released; Schema Intelligence preserves the mappings among action, authority, execution interface, and monitoring signal. Once released, the intervention enters established operating processes, where delay, deviation, override, interruption, failure, and recovery become structured execution evidence.

\noindent\textbf{5. Feedback Learning.} Feedback Learning turns execution evidence into structured revision of the decision context. The Site World updates represented state, feasibility conditions, and action-consequence estimates; Schema Intelligence links realized actions and observed outcomes back to the formulation, assumptions, models, authority conditions, and feedback signals; the Organizational World evaluates whether the course of action, trade-offs, commitments, or decision rights require authorized revision. This lineage supports diagnosis without assuming that every delayed or endogenous outcome can be causally assigned to the focal action. The enactive claim is therefore not simply that the system learns from outcomes, but that feedback can revise the relationships through which future AI-mediated action remains purposeful, feasible, accountable, and adaptive \citep{sterman1994learning}.

Across decision episodes, the cycle connects organizational and operational cadences. Routine decisions may proceed within envelopes set by existing policies and tactical plans, while exceptions can propagate upward when they reveal that assumptions, constraints, commitments, or rights are no longer tenable. In this sense, the cycle sustains situated decision coherence while preserving differentiated responsibilities across organizational levels.

\subsection{System-Level Validity Governance}

The evaluation and governance of Enactive AI are not separate concerns added after model use; they are the conditions under which the architecture remains valid while it is being used. Evaluation determines what the system treats as improvement, while evidentiary governance determines whether that judgement is grounded in traceable and current system evidence. If evaluation collapses to a narrow local indicator, the architecture may optimize a visible fragment while weakening broader purpose, feasibility, resilience, or accountability. If evidence lineage, freshness, semantic consistency, or authority records deteriorate, the world models may continue to generate recommendations while no longer representing the system on which action depends. The central problem is therefore maintaining system-level validity as a decision is framed, formulated, grounded, released, executed, and revised.

Evaluation in this setting must follow the relationships that the architecture is designed to preserve. \emph{Objective alignment} asks whether an action still serves the organizational purpose and admissible trade-offs represented by the Organizational World. \emph{Feasibility preservation} asks whether the action remains compatible with the operating state, resources, timing, process dependencies, and safety boundaries represented by the Site World. \emph{Consequence evaluability} asks whether expected consequences are represented in a form that allows later comparison rather than being compressed prematurely into an opaque score. \emph{Release integrity} and \emph{execution fidelity} ask whether the action was properly authorized and whether the commanded action remained aligned with the realized action. These criteria are not merely performance measures; they prevent the system from mistaking local metric improvement for system-level progress.

The same criteria require corresponding evidentiary governance. Objective alignment requires authoritative records of goals, commitments, policies, trade-offs, and decision rights. Feasibility preservation requires current operating-state evidence, resource availability, constraint status, uncertainty, and safety margins. Consequence evaluability requires versioned assumptions, model interfaces, candidate-action descriptions, and distinguishable consequence dimensions. Release integrity and execution fidelity require auditable records of recommendation, approval, release, commanded action, realized action, deviation, exception, interruption, failure, and recovery. Governance is therefore not an archival function. It is the evidentiary condition that allows evaluation to remain system-wise rather than drifting toward whichever indicator is easiest to observe.

The architecture distributes these validity-maintenance responsibilities across its components. Schema Intelligence maintains provenance, freshness, semantic consistency, versioned formulation and model lineage, uncertainty annotation, and the mappings among objectives, states, constraints, authority conditions, actions, and outcomes. The Site World maintains execution evidence and updates the operational representation when realized actions, state changes, deviations, or exceptions alter feasibility or consequence estimates. The Organizational World maintains the authoritative record of objectives, commitments, policies, decision rights, approval conditions, review responsibilities, and revisions. Together, these responsibilities do not guarantee full causal attribution in delayed or endogenous systems, but they make judgement and revision accountable to the evidence structure through which the decision was formed, released, executed, and learned from.

\subsection{Realization of Core Capabilities}

The three capability requirements identified earlier are realized by the architecture as coupled decision functions rather than as isolated model properties. Schema Intelligence supplies the semantic and relational substrate, the Organizational World reasons over purpose and coordinated courses of action, the Site World grounds action in operating conditions, and the Enactive Decision Cycle keeps these relations active across execution and feedback. This division of responsibility explains how the architecture supports system-grounded forecasting, consequence-grounded decision-making, and system-wise judgement within the same evolving decision system.

\noindent\textbf{System-grounded forecasting.} In Enactive AI, forecasting is a coupled projection of how a course of action may propagate through the system. Schema Intelligence supplies the semantic and relational structure that makes the relevant entities, dependencies, assumptions, model interfaces, and feedback links mutually legible. The Organizational World projects organizational consequences: whether the course remains coherent with purpose, commitments, authority, incentives, and cross-unit coordination. The Site World projects operational consequences: how the course interacts with current and anticipated states, resource constraints, process dependencies, execution windows, bottlenecks, and risk boundaries. Forecasting therefore rests on system modeling and coupling, not on data-to-outcome extrapolation alone; its purpose is to anticipate how an AI-mediated intervention may change the conditions under which the system subsequently acts.

\noindent\textbf{Consequence-grounded decision-making.} Consequence-grounded decision-making is realized by converting projected consequences into an intervention that can pass tests of feasibility, reliability, safety, and logical coherence. The Organizational World keeps objectives, commitments, trade-offs, accountability, and decision rights visible, so that the proposed action remains aligned with system-level purpose. Schema Intelligence preserves the mappings among objectives, states, constraints, authority conditions, models, and feedback, so that the reasoning chain remains traceable and internally consistent. The Site World evaluates whether candidate actions are technically feasible, execution-ready, risk-aware, and compatible with operating and safety constraints. A high-quality decision is therefore not simply the alternative with the highest local score; it is an intervention whose expected consequences are understood, whose release conditions are satisfied, and whose execution can be monitored and revised as the system responds.

\noindent\textbf{System-wise judgement.} System-wise judgement is realized inside the decision cycle through recurrent system determinations made by the two world models and connected by Schema Intelligence. The Organizational World judges whether the evolving course of action still preserves purpose, commitments, authority, accountability, and admissible trade-offs. The Site World judges whether the grounded intervention remains technically feasible, execution-ready, risk-bounded, and distinguishable in its projected and realized consequences. Schema Intelligence maintains the lineage among formulation, action, evidence, authority conditions, and feedback, so that these judgments refer to the same decision context. Judgement in this sense is not post-hoc scoring; it is the capacity to determine whether the cycle may proceed, must be revised, should be suspended, or requires authorized escalation as the system responds.

Together, these realizations clarify the architectural role of Enactive AI. The framework does not treat forecasting, action selection, and evaluation as separate stages connected only by handoff. It organizes them around a shared and revisable decision context in which purpose, state, feasible action, consequence, authority, execution, and feedback remain connected as the system changes.
\section{Industrial Illustrations of Enactive AI}
\subsection{JD.com: Connecting Consumer Demand with Supply-Chain Execution}
\label{sec:jd-longitudinal-illustration}

JD.com provides a longitudinal illustration of how Enactive AI can operate in a large-scale digital commerce and supply-chain setting. JD.com is a major Chinese retail and supply-chain service platform whose business combines online retail, marketplace services, and logistics-enabled fulfillment. It connects consumers with JD.com's self-operated retail units, third-party merchants, brand owners, and manufacturers, while also running warehousing, transportation, delivery, and after-sales capabilities that turn online transactions into physical service commitments. This dual role means that JD.com is not only an e-commerce interface for demand generation, but also an operator of a nationwide supply-chain execution system.

Its operational system must therefore coordinate commercial, inventory, and logistics decisions on a continuous basis. Assortment and procurement plans shape what products enter the system; demand forecasts inform where inventory should be placed; category and promotion decisions change substitution patterns and demand peaks; warehouse and transportation systems determine whether promised service levels can actually be met. This operational logic makes JD.com a hyper-scale complex system in which three dimensions of coupling are deeply intertwined. Process coupling arranges a continuous and mutually causal decision chain from assortment planning (which SKUs to introduce), demand forecasting (what sells where), category combination (what shares shelf space), inventory deployment (where to stock), to fulfillment scheduling (how to deliver fastest). A forecasting error does not remain local; it propagates into inventory misallocation, constrains warehouse throughput, degrades delivery lead times, and ultimately distorts the next cycle of assortment and forecasting. Organizational coupling involves sourcing teams pursuing margin and supplier terms, operations teams targeting turnover and cost reduction, and logistics teams committed to fulfillment rates and on-time delivery—each operating under KPI regimes that are inherently conflicting, such as maintaining high inventory for service reliability versus keeping low inventory for working capital efficiency. Ecosystem coupling further extends the boundary to third-party sellers whose inventory, pricing, and promotion decisions interact with JD.com’s platform rules and logistics capabilities. In such a tightly coupled system, any isolated optimization—however sophisticated its predictive model—risks being eroded by cross-functional feedback delays and objective misalignment, precisely the conditions that require a decision-centric architecture for complex systems.

For JD.com, the relevant question is therefore not only how to predict demand more accurately, but how to keep commercial intent, inventory decisions, and fulfillment execution aligned as conditions change. Enactive AI provides a way to organize this alignment through three connected roles. The Organizational World explicitly represents the decision logic, KPI weights, and authority boundaries of each business unit. It models the inherent conflicts—for instance, when a major promotion season arrives, the Organizational World generates a system-level course of action that adjudicates whether margin concessions are admissible to preserve fulfillment service, or whether increased safety stocks are authorized to reduce stockout risks, thereby defining which unit yields under which condition for global coherence. The Schema Intelligence addresses the fragmentation of JD.com’s legacy systems, where product master data, supply-chain records, warehouse management system (WMS) logs, transportation management system (TMS) traces, and merchant-platform data reside in heterogeneous architectures. Its role is to construct a cross-functional unified semantic network that maps product attributes to forecasting features, forecasting outputs to inventory-level constraints, and inventory positions to warehouse operational capacities. Only through this systematic integration can Enactive AI support globally coherent decisions rather than forcing incompatible data silos into a single model. The Site World, in JD.com’s context, bounds the physical layer of warehouse operations—receiving, putaway, picking, packing—and the dispatch optimization of line-haul and last-mile delivery networks. It provides a decision-sufficient representation of current operational reality: warehouse congestion levels, picker workload saturation, available vehicle capacity, and real-time traffic conditions. While the Organizational World determines what to deliver and where, the Site World qualifies whether and how it can be executed reliably under current physical constraints.

Our joint technical work with JD.com shows how Enactive AI takes shape through concrete operational capabilities. System-grounded forecasting begins with models of new-product life-cycle curves and product attributes through deep schema modeling, embedding market trends, promotional calendars, and seasonality into the forecasting engine beyond mere statistical fitting \citep{lei2023lifecycle,lei2025pooling}. It is further extended by TimeHF, a 6-billion-parameter time-series model guided by human feedback that supports automated replenishment for more than 20,000 products in JD.com's supply chain and reports a 33.21\% accuracy improvement over existing methods \citep{qi2025timehf}. Consequence-grounded decision-making materializes in end-to-end inventory management and assortment studies, where physical constraints from the Site World and financial constraints from the Organizational World are jointly embedded into the decision model—ensuring that recommended inventory policies are not only mathematically optimal but also organizationally admissible and physically executable \citep{qi2023e2e,liu2026transformer}. System-wise judgement is exemplified by the integrated assortment and inventory-allocation system across the two-level RDC-FDC network, which performs the Enactive Decision Cycle in a concrete fulfillment setting: framing headquarters’ service targets (Intent Framing), grounding decisions in network-wide inventory and available capacity (State Grounding), evaluating fulfillment and transfer actions (Action Evaluation), executing through established logistics processes (Operational Enactment), and revising representations based on realized sales and replenishment delays (Feedback Learning) \citep{shen2025fulfillment,hu2026warehouse}. Collectively, these examples show how the forecasting, optimization, and execution modules of Enactive AI can be implemented at varying granularities across specific JD.com tasks.

Deploying these Enactive AI methodologies across JD.com’s massive operational footprint has produced not merely local efficiency gains, but a systemic shift from reactive fulfillment toward proactive, coordinated supply-chain intelligence. The COVID-19 resilience case reported by Shen and Sun offers a compact stress test: when the Wuhan RDC lockdown disrupted Hubei distribution, JD.com adapted JD-NetSIM and delivery procedures to reroute demand, support contactless delivery, and coordinate emergency supplies with Hubei authorities, with the paper reporting better 48-hour delivery performance and lower cost than a manual emergency plan \citep{shen2021covidresilience}. The economic and operational outcomes are substantial: the integrated network deployment achieved approximately \$6.13 million in annual FDC holding and capital cost savings and \$22.32 million in annual transfer cost reductions \citep{shen2025fulfillment}. In a rigorous 30-day matched field experiment covering 61,430 orders, 12 distribution centers, and 9,308 SKUs, the system reduced holding costs by 26.1\%, stockout costs by 51.7\%, and total inventory costs by 40.4\% relative to JD.com’s prior practice \citep{qi2023e2e}. At the service level, full deployment across all eight RDCs and their subordinate FDCs raised stock availability by 0.85 percentage points and local fulfillment by 2.19 percentage points, with an estimated 18.61 million orders annually benefiting from improved delivery timeliness or customer experience \citep{shen2025fulfillment}. This body of work has also received broad industry and academic recognition: the integrated assortment and inventory-allocation project received the 2024 Daniel H. Wagner Prize \citep{informs2024wagner}; JD.com’s broader supply-chain program was named a 2023 Franz Edelman Award finalist \citep{informs2023edelman}; a demand-prediction study was a finalist in M\&SOM’s 2023 practice-based research competition \citep{informs2023practice}; and JD.com was awarded the 2024 INFORMS Prize at the organizational level \citep{informs2024prize}. These validated outcomes, drawn from field experiments, deployment reports, and aggregate operational metrics, illustrate how the three architectural roles of Enactive AI can remain distinct yet tightly connected across repeated decision cycles. They provide a concrete bridge from the architecture to the paper’s later evaluation agenda; causal validation of the complete Enactive AI framework in its full generality remains an open empirical task, to which the JD.com record serves as a foundational existence proof.

\subsection{Enactive AI in a Leading Telecommunications Operator}
A leading telecommunications operator in China illustrates cross-level decision-making in a supply network of extraordinary scale. The operator continually builds and upgrades base stations, core-network nodes, and transmission infrastructure across regions, translating annual network plans into equipment procurement, on-site installation, and project-delivery tasks. Critical spare parts must reach maintenance sites in time to prevent local equipment failures from escalating into service disruptions, while mobile phones, routers, and other consumer devices must be replenished and reallocated between regional warehouses and retail service outlets as local demand changes. The stability of this supply chain affects national information infrastructure, business continuity, and everyday communication services. The case therefore centers on how Enactive AI organizes decisions that connect infrastructure investment, service continuity, procurement, inventory, logistics, and consumer-facing availability.

In recent years, geopolitical disruption, trade friction, and rapid technological change have accelerated the fragmentation and reconfiguration of global supply chains. Internally, the operator manages a supply system whose complexity arises from the interaction among material structure, supplier ecology, and spatial execution.

The first source of complexity is the multilayered product structure. Telecommunications operations involve base-station equipment, core-network systems, transmission equipment, optical cables, spare parts, consumer devices, and general supplies. A single piece of equipment may contain many modules and lower-level components, while the same critical component may be used across several equipment models. Technology upgrades alter compatibility requirements, spare-parts demand, and life-cycle costs. When a supply interruption, equipment upgrade, or new procurement cycle occurs, the relevant decision unit is therefore not an isolated SKU but an equipment-module-component configuration over its life cycle.

A second source of complexity is the ecosystem of thousands of business partners whose risks are not independent. Although the operator contracts mainly with first-tier equipment and product suppliers, several of them may depend on the same second-tier manufacturer, chip producer, raw-material source, or regional production node. Procurement sources that look diversified at the contractual level may therefore share a common upstream exposure. First-tier performance records alone cannot reveal which equipment, orders, regions, and service commitments will be affected by an upstream event.

The third source of complexity is spatial execution. Materials move across supplier tiers, administrative regions, warehouses, and logistics nodes, from cross-regional hubs to provincial, municipal, and county-level warehouses. Network equipment may be shipped to base stations, equipment rooms, transmission nodes, or installation sites; spare parts support maintenance and fault recovery; and consumer devices must reach retail outlets in different cities. Moving inventory to one region reduces its availability elsewhere, while prioritizing network construction may consume logistics capacity needed for consumer devices.

The difficulty is therefore not scale alone, but tight coupling. Existing systems may separately support demand forecasting, procurement management, supplier evaluation, inventory visibility, warehouse operations, transportation planning, and network-project management. These capabilities often operate with different data, performance measures, and business boundaries. Applying similar management logic to strategic equipment and high-volume terminal products may leave critical components insufficiently traceable while creating excessive inventory for standardized items. Supply-risk visibility may stop at first-tier suppliers, and locally defined logistics rules may impede cross-regional resource allocation.

Enactive AI addresses these problems by keeping decision purpose, supply relationships, operational feasibility, and authorized execution connected. The \emph{Organizational World} captures the higher-level concerns and authority boundaries that shape supply-chain decisions. At the corporate level, the operator must balance network continuity, supply security, and investment efficiency, while provincial and municipal subsidiaries remain responsible for local construction, maintenance, and service commitments. Network-infrastructure teams may reserve critical equipment for construction projects; consumer-facing units need device availability at retail outlets; procurement and finance control purchasing, emergency payment, and contractual compliance; and corporate approval may be required for cross-provincial transfers or major project delays.

\emph{Schema Intelligence} connects the organizational, supply, and operational levels by maintaining the relationships through which decisions propagate. It aligns equipment models and bills of materials with module and component codes, actual manufacturers, contracted suppliers, and upstream and downstream supply links, allowing hidden dependencies to be identified across otherwise separate information systems. It also represents spatial and fulfillment relationships within the warehouse network. For example, it records where a particular batch is stored, whether it has already been reserved for a construction project, which logistics routes can deliver it to an installation site, and whether replenishing a retail outlet would compete with network equipment for warehouse or transport capacity. In this way, an objective such as “protect network continuity in priority regions” can be translated into specific projects, equipment, spare-parts levels, and delivery deadlines, while a supply-risk signal becomes a list of affected orders, inventories, projects, and actions requiring reevaluation.

The \emph{Site World} represents the bounded operational environment in which candidate actions take place. In this case, it links telecommunications-network sites, supply and inventory sites, and consumer-service sites: base stations and equipment rooms, construction and maintenance tasks, warehouses, available and reserved inventory, materials in transit, supplier capacity, transport routes, retail outlets, local demand, and replenishment status. It does not reproduce the operator's entire operating environment, but retains the information that can materially change risk exposure, action feasibility, execution acceptance, or expected consequences.

The Enactive Decision Cycle provides a common setting in which these operational issues can be addressed together. Consider a disruption in which a geopolitical event or natural disaster reduces the capacity of a second-tier supplier whose critical component is used by several first-tier equipment suppliers. The disruption affects multiple equipment models, construction projects, and maintenance requirements. During \emph{Intent Framing}, decision-makers establish which network projects, service commitments, and retail operations receive priority, together with acceptable cost, delay, and authority boundaries. They can then consider interventions such as cross-regional inventory transfers, expedited transportation, alternative sourcing, supplier-capacity reallocation, equipment substitution, procurement-plan adjustments, or postponement of lower-priority projects.

During \emph{State Grounding}, Schema Intelligence identifies the equipment, outstanding orders, and business tasks associated with the affected materials. The Site World then establishes which warehouse inventories are genuinely available, which have already been committed to projects, and how long current resources can sustain the affected operations. The system distinguishes confirmed production stoppages from unverified delivery delays and potential exposures supported by incomplete evidence. The result is not simply a supplier risk score but a decision-specific view of which commitments are at risk, when the effects are likely to emerge, and which parts of the assessment remain uncertain.

During \emph{Action Evaluation}, the system compares cross-functional responses. Inventory reserved for a project that has not yet entered its construction window might be transferred to a priority region facing a shortage of maintenance spares, with the original project rescheduled accordingly. Suppliers might reallocate limited capacity and use expedited transportation. Certified substitute components could be introduced for eligible equipment models, allowing the sequence of equipment upgrades to be revised. Each option is assessed in terms of effects on maintenance capacity in other regions, project schedules, consumer-device replenishment, procurement cost, and future supplier dependence. A cross-provincial transfer, for example, may protect a priority project while leaving the sending region close to its minimum spare-parts threshold. Technical certification, contractual conditions, transport capacity, construction windows, and approval requirements remove options that cannot be executed.

During \emph{Operational Enactment}, the selected response is authorized and translated into established business processes. A cross-provincial transfer requires confirmation from the relevant provincial subsidiaries and corporate functions; a substitute component requires technical validation; and emergency procurement must pass contractual, financial, and compliance review. Execution may nevertheless deviate from the approved plan because inventory is frozen, transportation is disrupted, or supplier capacity changes, so the system distinguishes among recommended, approved, commanded, and realized actions. During \emph{Feedback Learning}, actual deliveries, project use, and regional inventory changes update supplier-performance assessments, logistics lead times, and action feasibility, while also revealing whether the intervention has transferred risk to another region.

The system operates across several time scales. The fast-response layer manages inventory allocation, transportation, and project adjustments during an active disruption. At an intermediate horizon, recurring anomalies inform revisions to safety stocks, supplier allocations, and material-criticality classifications. At a slower strategic horizon, the organization reassesses supplier diversification, equipment technology pathways, warehouse-network design, and corporate governance rules. The value of Enactive AI lies in connecting these time scales through a shared decision history: it supports strategic asset life-cycle management, network-risk and supplier-ecosystem governance, coordinated procurement-inventory-logistics operations, and compliance-aware human--AI collaboration without allowing short-term feedback to modify long-term policy without appropriate authorization.

\subsection{AMHS Decision-Making: Enacting a Site World in Semiconductor Manufacturing}
\label{sec:amhs-site-world}

Automated material handling systems (AMHS) in semiconductor manufacturing provide a mechanism-revealing setting for Enactive AI. Overhead hoist transport (OHT) vehicles move front-opening unified pods among tools, stockers, and buffers through a shared rail network. AMHS operation is tightly coupled with fab production: a delayed or poorly prioritized movement can starve an expensive tool, while a locally attractive dispatch or route can shift congestion to another segment, delay later tasks, or become infeasible before release \citep{agrawal2006survey,sun2005integration}. 
These properties make AMHS more than a routing problem. The network topology is explicit, operating states change continuously, actions alter the state faced by other vehicles, safety constraints are non-negotiable, and execution generates fine-grained event traces.

Existing research has separately improved vehicle allocation, congestion-aware routing, predictive empty-vehicle management, graph-based traffic prediction, and layout-performance analysis \citep{lin2013dynamic,bartlett2014congestion,schmaler2017empty, hwang2020qroute,wu2021performance,ahn2022congestion,hong2022practical,chou2025multiagent}. 
This literature establishes that dynamic dispatching, reinforcement learning, graph models, and simulation are already important AMHS capabilities. The architectural claim here is therefore not that Enactive AI supplies another routing algorithm. It is that a bounded operational representation can keep production purpose, current state, feasible intervention, release authority, physical execution, and subsequent evidence connected even when different analytical and control mechanisms remain modular.

The \emph{focal decision} is a recurrent transport intervention. Given a material-movement request and the current operating conditions, the system must determine which eligible OHT should serve the request, which feasible route or next hop it should take, and whether the movement should be released or held. This definition distinguishes the decision from the methods used to support it. A shortest-path routine, learned value function, traffic predictor, optimization model, or discrete-event simulation may generate or evaluate candidates, but none of these mechanisms alone determines the decision's production objective, admissible scope, release authority, or connection to realized execution. It also separates decision horizons. Predictive repositioning of empty vehicles changes the fleet supply available to later transport episodes, while track-layout and shortcut decisions establish a slower-changing topology and capacity envelope. They are coupled to the focal decision without being treated as the same action.

For this focal decision, the \emph{Site World} is a decision-sufficient representation rather than a comprehensive digital replica of the fab. Its state includes the directed rail topology; vehicle locations, directions, speeds, loads, availability, health, and remaining assignments; transport-request origins, destinations, priorities, and timing; FOUP and process status; track occupancy, local queues, intersection contention, observed travel times, and congestion; stocker, buffer, load-port, and equipment readiness; blocked segments, regional capacity, alarms, and maintenance restrictions; and the provenance, freshness, and uncertainty of these observations. A small state difference must be retained when it changes reachability, safety, controller acceptance, or the expected consequences of a candidate action. Details that do not materially change the relevant action set or its consequences need not enter the representation.

Candidate actions include vehicle assignment, route or next-hop selection, release, temporary holding, rerouting, and bounded fallback. Each candidate is related to differentiated consequences such as pickup waiting, delivery time, downstream tool starvation, congestion propagation, empty travel, fleet utilization, recovery requirements, and uncertainty. Technical qualification requires compatibility with directed reachability, segment and regional capacity, mutual exclusion, vehicle separation, loading and unloading availability, timing requirements, controller interfaces, and independent safety interlocks. The Site World can therefore establish that an intervention is technically executable, but it does not independently decide that the intervention is organizationally admissible. Nor does it replace the material control system (MCS), OHT controller, deterministic traffic control, or safety mechanisms that may reject, interrupt, or override a released action.

The \emph{Organizational World} supplies the slower-changing managerial envelope within which these high-frequency decisions acquire meaning. Relevant elements include fab cycle-time and throughput objectives, commitments to protect critical tools or high-priority lots, service levels for ordinary and urgent transports, acceptable congestion and recovery risk, escalation conditions, and the rights to modify priorities, release an intervention, suspend automation, or revise a policy. This arrangement does not imply managerial approval of every millisecond- or second-level routing choice. Routine actions may be released automatically within a preauthorized decision envelope; unfamiliar states, weak evidence, conflicting production priorities, excessive consequence severity, or proximity to a safety boundary trigger fallback or escalation.

\emph{Schema Intelligence} connects this managerial envelope to the physical representation. It links MES lots, process steps, equipment states, and production priorities to MCS transport jobs; connects those jobs to OHT-controller vehicles, nodes, edges, and commands; associates safety and maintenance policies with admissibility and feasibility conditions; and preserves relationships among recommendations, approvals, commands, realized trajectories, and outcomes. For example, the objective of avoiding starvation at a critical tool becomes operational only when it is connected to expected process completion, the affected FOUP and transport request, an applicable time window, available vehicles and routes, and the authority to override an incumbent priority. These decision-specific mappings provide objective-to-measure lineage, operating-state-to-action feasibility, decision-right-to-release congruence, and action-to-outcome traceability. They also preserve distinctions that generic data integration can obscure: a technically feasible route is not necessarily authorized, a recommendation is not a command, and a command is not identical to the movement ultimately realized.

The five stages of the \emph{Enactive Decision Cycle} show how these representations become operational. During \emph{Intent Framing}, the responsible owner specifies the production commitment to protect, decision scope and horizon, performance measures, acceptable trade-offs, escalation conditions, and authority boundary. Measures such as transport time, tool starvation, congestion, fleet use, and recovery risk remain distinct rather than being prematurely collapsed into one score. During \emph{State Grounding}, Schema Intelligence links that mandate to current MES, MCS, controller, equipment, and track evidence. The Site World determines whether the evidence is sufficiently current and informative; missing events, stale queue estimates, topology changes, or disagreement among systems become decision-relevant uncertainty.

During \emph{Action Evaluation}, modular analytical mechanisms generate technically qualified vehicle--route--release candidates and estimate their differentiated consequences. A routing model may compare next-hop values using destination, topology, queue, congestion, and observed travel time; a vehicle-assignment model may include both idle vehicles and vehicles nearing task completion; a demand or heat-map model may estimate future regional supply gaps; and simulation may expose congestion propagation and resource contention under controlled counterfactuals. The cycle requires alternatives to be compared against the focal production objectives and constraints, rather than automatically selecting the lowest predicted travel time.

During \emph{Operational Enactment}, feasibility is revalidated against the latest state immediately before release. The relevant decision-right holder or governed workflow determines whether the technically qualified intervention is admissible. Once released through established MCS and OHT-control interfaces, independent interlocks and traffic-control mechanisms remain authoritative. The Site World records the recommended, approved, commanded, and realized actions separately, together with delay, override, rejection, interruption, fallback, and recovery. During \emph{Feedback Learning}, expected consequences are compared with realized movement, waiting, travel time, queue formation, congestion, tool-service effects, and execution exceptions. This evidence may revise state estimates, transition representations, candidate-generation rules, or the mappings maintained by Schema Intelligence. It supports diagnosis and traceability, but does not by itself establish that the focal recommendation caused an observed production outcome. Changes to production priorities, risk tolerances, or release rights require an authorized organizational process.

This cycle operates at two connected speeds. A fast inner loop updates the operational state, evaluates vehicle--route--release candidates, issues a governed command, and incorporates execution events. A slower outer loop reviews repeated exceptions, model drift, threshold adequacy, cross-area spillovers, and the continued validity of the preauthorized decision envelope. Empty-vehicle repositioning can operate between these loops by using predicted regional demand and available supply to shape the future state. Layout, shortcut, and regional-capacity decisions operate more slowly still, using accumulated evidence and counterfactual simulation to alter the structural environment within which online decisions are made. This nested structure turns the project's proposed line from passive response to anticipatory operation into a governed cross-horizon mechanism rather than unconstrained online self-modification.

The available project artifacts provide design and model-validation evidence for this instantiation, not field validation of the complete Enactive AI architecture. The technical design specifies a closed loop from task generation and event-driven simulation through unified vehicle, track, task, regional, and congestion states; demand and congestion prediction; routing and empty-vehicle decisions; regional admission and deadlock prevention; execution feedback; and longer-horizon layout evaluation. The implementation blueprint connects online MES, MCS, and OHT-controller data to world modeling, policy evaluation, execution, hard constraints, fallback, historical replay, stress testing, and staged shadow operation. These artifacts make the intended Site World and enactment interfaces concrete.

The evidence boundary must remain explicit. An internal simulator-calibration study uses a real rail map and historical tasks from a seven-OHT setting, with ten replications per strategy, and reports a put-down-path exact-match rate above 99\% together with broadly aligned transport-time and vehicle-state measures. This evidence supports grounding of the simulator within the tested setting; it does not demonstrate the performance of the complete architecture. The same report describes component-level simulation results, including an improvement of up to approximately 8.8\% in average task-completion time under a medium-load condition and approximately 22--23\% more completed tasks during an initial warm-start window relative to a cold start. These results are model-contingent and should not be interpreted as causal improvements in fab throughput, cycle time, yield, or reliability. Planned simulation, shadow, and pilot stages are likewise an evaluation design rather than realized deployment evidence.

A prospective evaluation should compare four configurations under common state data, action candidates, hard constraints, and computational budgets: the incumbent MCS/OHTC policy; a strong dynamic dispatching and routing baseline; a Site-only analytical configuration that predicts congestion and optimizes vehicle--route choices without production-purpose mappings or complete execution lineage; and the full Enactive configuration with objective mappings, preauthorized release boundaries, commanded-to-realized traceability, and governed feedback revision. Component ablations can then test whether uncertainty representation, Schema mappings, execution lineage, and the slower review loop account for incremental value beyond model capacity.

Validation should proceed from independent-period simulator calibration and stress tests, through frozen-policy shadow operation, to a bounded field pilot. Shadow operation can measure candidate feasibility, controller rejection, fallback, latency, and disagreement with the incumbent policy, but cannot establish production benefit because the recommendations are not enacted. Where safety and operational conditions permit, a field evaluation could use time-block or weakly coupled subnetwork switchbacks with an appropriate washout period; randomization by individual transport request would ignore interference created by shared vehicles and congestion. Primary outcomes should include transport-related tool starvation or blocking, hot-lot tardiness, regional or fab cycle time, and throughput. Secondary outcomes include pickup and delivery-time distributions, severe-congestion frequency and recovery, empty travel, utilization, infeasible recommendations, controller rejection, overrides, decision latency, lineage completeness, human workload, and the cost of maintaining the Site World and its mappings.

AMHS thus illustrates the distinctive architectural claim of Enactive AI. Decision intelligence does not arise merely from inserting a stronger routing model into an existing controller. It arises from preserving operational decision coherence as organizational purpose is translated into a bounded physical state, candidate actions are technically qualified and organizationally released, execution changes the shared environment, and realized consequences return as governed evidence. The Site World provides the bounded physical grounding; Schema Intelligence preserves cross-system meaning and lineage; and the Enactive Decision Cycle keeps the representation revisable without conflating simulation, execution, safety, and causal learning.

\section{Research Agenda}
\label{sec:research-agenda}

Enactive AI reframes the future research agenda for AI systems that act in complex organizational and operational environments. The agenda is not simply to improve isolated model capabilities, automate more workflows, or attach governance mechanisms after deployment. The deeper shift is in the unit of inquiry. Once AI is understood as part of an evolving decision system, research must examine how purpose, evidence, constraint, action, consequence, authority, execution, and learning can remain connected as the system changes. We therefore highlight six research directions that are central to whether Enactive AI can participate deeply in system-level decision-making and produce beneficial, accountable, and adaptive effects in the systems where it is used.

These directions form two connected streams. The first three constitute a technical agenda for building Enactive AI systems that can understand, enact, and adapt within complex systems. They concern the accuracy of system representation, the consistency of the decision-to-execution process, and the resilience of the AI decision architecture under change. The latter three constitute a managerial agenda for governing, evaluating, and legitimizing Enactive AI in organizations and society. They concern organizational design and responsibility, system-level value assessment, and the societal value conditions under which AI-mediated decision systems can be accepted and sustained.

\subsection{Technical Agenda}

The first stream concerns the technical foundations required for Enactive AI to operate in complex systems. Its focus is not a single model capability, but the construction of a decision architecture that can sense and represent the system, maintain consistency as decisions move toward execution, and continue operating as the system changes.

\subsubsection{Decision-Oriented System Representation and Sensing}

Future research should investigate how AI can form system understanding that is sufficient for situated action. This is not only a question of what information should be represented, but also a question of how heterogeneous system information should be organized so that AI can reason about system evolution, cross-component relationships, and the consequences of intervention. Enactive AI therefore requires research on representational structures that connect sensed data, organizational records, operating states, resource conditions, constraints, action spaces, and feedback signals into forms usable for decision-making rather than merely for description.

This direction requires moving beyond comprehensive representation as an ideal. In complex systems, no AI system can or should represent everything. The research problem is how to organize decision-relevant information into structures that preserve the relations needed for action: objective-to-measure relations, state-transition relations, resource and constraint dependencies, action-consequence pathways, authority conditions, feedback loops, and evidence lineage. Future work may therefore study how Enactive AI identifies which structures are needed for a focal decision, how these structures should combine symbolic, statistical, operational, and organizational information, and how sensing and evidence pipelines should determine whether the system representation remains current enough for action.

\subsubsection{Reliable Enactment and Process Consistency}

Future research should study how Enactive AI can move from system understanding to reliable enactment. The central problem is not only whether AI can recommend a plausible action, but whether the decision process can remain consistent as an intention is translated into a formulation, grounded in current conditions, evaluated for feasibility, mediated through human interaction, and released into execution. There is still no settled science of what constitutes a good AI-mediated execution process in complex systems, where feasibility, authority, timing, human response, and operating constraints may all change while the decision is being prepared and enacted.

This direction also responds to problems already visible in current AI applications. Systems often struggle with context selection, retrieval relevance, summarization loss, source grounding, and hallucinated or weakly supported claims. These failures can shift what the task appears to be, which evidence is treated as salient, and which objective the system implicitly pursues. In complex-system decision-making, the same problem becomes more consequential because the relevant context is distributed across organizational purposes, operating states, policies, constraints, responsibilities, and feedback. A small distortion in context may redirect objectives, hide trade-offs, narrow authority conditions, or convert a system-level concern into a convenient local task.

Future work may therefore develop mechanisms for maintaining process consistency across the decision cycle. Such mechanisms could test objective fidelity, semantic consistency, evidence provenance, authority completeness, feasibility preservation, human override conditions, and sensitivity to missing assumptions before a formulation is used for action evaluation or execution. They could also detect when a local proxy has displaced the system-level purpose, when conflicting evidence requires reframing rather than continued optimization, and when exceptions or execution deviations reveal that the decision has been formulated around the wrong boundary. Reliable enactment is therefore not a matter of automating execution alone; it is the capacity to keep intention, evidence, feasibility, human interaction, and realized action aligned while the system is moving.

\subsubsection{Adaptive Resilience of the AI Decision Architecture}

Future research should explore how Enactive AI can remain stable and useful as the systems around it change. Much current discussion of AI adaptation remains centered on updating model parameters, extending memory, refreshing external documents, or improving behavioral policies. These forms of adaptation are important, but they may be insufficient for AI systems that participate in system-level decision-making. In real operational settings, data definitions may change, factories may be reconfigured, business priorities may shift, suppliers may disappear, and employees with tacit knowledge may leave. The question is how an AI decision system can absorb such changes incrementally while continuing to operate, rather than requiring full retraining, complete redesign, or repeated reimplementation.

This direction should study robustness and resilience as conditions for maintaining stable operation rather than as isolated model properties. Future work should examine how Enactive AI preserves a workable decision state under expected variation, unfamiliar disruption, structural change, and unknown risks. This requires architecture-level adaptation: when the boundaries of the Site World should be expanded or narrowed, when Schema Intelligence must introduce new mappings or retire obsolete ones, when organizational decision rights should be reallocated, and when feedback channels or evaluation criteria must be redesigned. Consequences from real decision episodes are not merely additional training signals; they are evidence about whether the AI decision architecture still matches the system it is helping to shape.

\subsection{Managerial and Societal Agenda}

The second stream concerns the managerial and societal conditions under which Enactive AI can be governed, evaluated, and accepted. Once AI systems participate in consequential decisions, their design cannot be separated from organizational forms, responsibility structures, value assessment, and the social boundaries that define legitimate action.

\subsubsection{Human--AI Organization, Governance, and Responsibility}

Future research should study how organizations are reconfigured when AI becomes capable of participating in decision processes. This is not merely a question of keeping humans in the loop or assigning a final approver to an automated recommendation. Enactive AI suggests that organization itself may become a site of innovation: decision routines, role boundaries, coordination mechanisms, escalation paths, audit structures, and learning processes may need to be redesigned when AI systems continuously structure context, evaluate actions, monitor execution, and feed consequences back into future decisions.

This direction calls for a deeper account of human--AI organization as a new form of distributed decision capacity. Human actors may define purposes, adjudicate value conflicts, challenge evidence, authorize boundary changes, and take responsibility for institutional commitments. AI systems may preserve decision context, surface dependencies, compare feasible interventions, detect exceptions, and support cross-unit coordination. The research problem is how these capacities should be combined into organizational forms that are more than human decision-making with AI assistance or automation with human oversight. Future work should examine whether Enactive AI enables new patterns of hybrid expertise, cross-level coordination, and collective judgment in which humans and AI systems jointly sustain decision coherence over time.

Decision rights and auditability should be studied as part of this broader organizational architecture. Future work should examine how organizations determine which decisions can remain within preauthorized envelopes, which require human approval, which require escalation, and which should be withheld because evidence, authority, or consequence estimates are insufficient. It should also study how responsibility is preserved across the chain from recommendation to approval, command, realized action, and review. This includes incentive questions that may appear as governance or ethics problems on the surface: departments may resist a unified organizational model, employees may disagree with how their knowledge is codified, and local units may have reasons to withhold, reinterpret, or contest AI-generated system views. Governance in Enactive AI is therefore not only control over AI behavior; it is the design of organizational conditions under which AI-mediated action can remain reliable, contestable, and accountable.

\subsubsection{Evaluation of System-Level Value and Decision Contribution}

Future research should develop evaluation methods for AI-mediated decision systems. If AI is part of an evolving decision system, it can no longer be evaluated only as a technical performer. Predictive accuracy, benchmark scores, objective values, and task completion remain useful, but they do not answer the central question raised by Enactive AI: whether AI improves the coupled decision system in which action takes effect. Evaluation must therefore move from local performance to system-level contribution.

This direction should study how organizations and researchers can assess net system-level value. Relevant outcomes include objective alignment, feasibility preservation, consequence evaluability, release integrity, execution fidelity, resilience, coordination quality, human workload, accountability, adaptability, and traceability. They also include the cost of constructing and maintaining the decision context through which these properties are achieved. This agenda calls for empirical designs that can capture delayed effects, spillovers, execution gaps, human response, cross-cycle learning, and the possibility that a local gain weakens the broader system. A mature evaluation science for Enactive AI would make it possible to compare not only models or policies, but different architectures for sustaining reliable and valuable action in complex systems.

\subsubsection{Societal Values, Legitimacy, and Acceptance}

Future research should examine the societal value conditions under which Enactive AI can legitimately participate in consequential decision-making. This direction should not be treated as a peripheral ethical add-on or as a problem of organizational efficiency. When AI systems help allocate resources, prioritize risks, structure opportunities, govern work, or shape public and economic life, they encounter social boundaries that cannot be reduced to system performance. Human agency, dignity, privacy, non-discrimination, procedural justice, rights to explanation and appeal, and limits on delegating judgment become part of the conditions under which AI-mediated action can be authorized.

This agenda includes questions that reach beneath particular organizational frictions. Which decisions should remain non-delegable because they involve human rights, moral responsibility, or public authority? How should affected individuals and communities contest the evidence, objectives, or consequences represented by an Enactive AI system? How should historical inequality, biased institutional data, or uneven capacity to benefit from AI-supported decisions be addressed before action is released? Future work should therefore study ethical and institutional safeguards such as participation, explanation, appeal, contestability, proportionality, and independent review. The purpose is not only to avoid harm, but to clarify the social red lines and value commitments that must bound AI action in human systems.

Taken together, these research directions define Enactive AI as both a technical and managerial research program. The technical agenda asks how AI can represent, enact, and adapt in systems whose states, constraints, and responses evolve. The managerial agenda asks how organizations and societies can govern, evaluate, and legitimize such participation so that AI-mediated action remains valuable and accountable. A further research challenge lies in the coupling between these two streams: technical adaptation and managerial governance cannot evolve independently. A change in the Site World boundary may introduce new data sources and inference risks, while a change in decision rights may require the technical architecture to narrow autonomy, increase explanations, or alter feedback channels. Conversely, technical signals such as repeated exceptions, stale mappings, or uncertainty in execution evidence may need to trigger managerial responses such as authority reallocation, review procedures, or revised accountability rules.

This agenda points to a broader transformation in what AI research may become once intelligent systems are expected to act within complex systems rather than merely produce outputs about them. The long-term significance of Enactive AI lies not only in making AI more capable, but in making AI action more governable, more corrigible, and more durable in the systems where social, industrial, and economic outcomes are produced. It reframes future AI development around responsible system intervention: the ability to preserve purpose, remain grounded in changing conditions, adapt with the systems it helps to shape, and keep the consequences of action open to accountable revision.

\section{Conclusion}
\label{sec:conclusion}

This paper has examined AI-mediated decision-making in complex systems and proposed \emph{Enactive AI} as a decision-centric framework for this emerging frontier. The argument began from the recent movement of AI toward action-bearing systems, situated this movement within the long-standing challenges of complex systems, and identified three requirements for reliable decision intelligence: system-grounded forecasting, consequence-grounded decision-making, and system-wise judgement. Building on these requirements, the paper developed Enactive AI as an architecture for keeping intent, evidence, state, feasible action, consequence, execution, responsibility, and feedback connected as decisions unfold, and used supply-chain, strategic-material, and cyber-physical manufacturing settings to illustrate the framework's relevance.

The timeliness of Enactive AI follows from the widening gap between AI capability and the scale of the systems into which that capability is being introduced. As AI enters organizations, infrastructures, and industrial networks, its significance is increasingly measured by how it reshapes relations among people, processes, resources, and institutions. The rise of action-capable AI therefore calls for an expanded horizon of AI research, one that moves from model performance toward system participation. Enactive AI provides a language for this broader horizon by placing decision-making, consequence, governance, and learning within the same developmental arc.

Seen from this broader horizon, the next important transformation of AI will occur through its relationship with complex systems. The systems that shape production, logistics, infrastructure, public service, and organizational coordination are dynamic, interdependent, and institutionally governed. AI's influence in these settings will grow through its ability to help such systems perceive changing conditions, reason about possible futures, act under constraints, and learn from the consequences of intervention. Enactive AI gives direction to this transformation by treating intelligence as something sustained across cycles of situated action.

This orientation also clarifies the research agenda that follows from the framework. Future work can be organized around a new class of questions: how to build AI systems that remain meaningful across organizational levels, how to evaluate contributions that unfold over time, how to embed accountability within adaptive action, and how to design autonomy that matures with evidence, trust, and governance. Taken together, these questions give Enactive AI its forward-looking significance and turn complex systems into a central arena for defining the next stage of AI.

By proposing Enactive AI, we aim to extend the horizon of AI development toward the larger complex systems in which intelligence acquires consequence. This perspective invites future research to treat AI as a participant in the organization, governance, and continuous improvement of real systems. Developed in this direction, AI may exert a deeper influence on human society, technological progress, and system governance by helping organizations understand complexity, act with greater responsibility, and learn from the futures their actions create. The promise of Enactive AI is therefore to orient AI toward a broader form of decision intelligence, one that can contribute to the long-term evolution of economic production, technological infrastructure, organizational coordination, and system governance.

\begingroup
\small
\bibliographystyle{plainnat}
\bibliography{references}
\endgroup

\end{document}